\documentclass[10pt,twocolumn,letterpaper]{article}

\usepackage{cvpr}
\usepackage{times}
\usepackage{epsfig}
\usepackage{graphicx}
\usepackage{amsmath}
\usepackage{amssymb}
\usepackage{booktabs}

\usepackage[pagebackref=true,breaklinks=true,letterpaper=true,colorlinks,bookmarks=false]{hyperref}

\usepackage{placeins}
\usepackage{multirow}
\usepackage{boldline}

\graphicspath{ {figure/} }

\cvprfinalcopy 

\def\cvprPaperID{****} 

\ifcvprfinal\fi
\begin{document}

\title{Weaving Multi-scale Context for Single Shot Detector}

\author{Yunpeng Chen $^{1}$ \quad Jianshu Li $^{1}$ \quad Bin Zhou $^{1}$ \\
\smallskip\smallskip\smallskip
Jiashi Feng $^{1}$ \quad Shuicheng Yan $^{2,1}$\\
\smallskip\smallskip\smallskip\smallskip
$^{1}$ National University of Singapore  \quad  $^{2}$ Qihoo 360 AI Institute \\
{\tt\small \{chenyunpeng, jianshu, bin.zhou\}@u.nus.edu \quad \{elefjia, eleyans\}@nus.edu.sg}
}

\maketitle

\begin{abstract}
   Aggregating context information from multiple scales has been proved to be effective for improving accuracy of Single Shot Detectors (SSDs)  on object detection. However, existing multi-scale context fusion techniques are computationally expensive, which unfavorably diminishes the advantageous speed of SSD. In this work, we propose a novel network topology, called WeaveNet, that can efficiently fuse multi-scale information and boost the detection accuracy with negligible extra cost. The proposed WeaveNet iteratively weaves context information from adjacent scales together to enable more sophisticated context reasoning while maintaining fast speed. Built by stacking light-weight blocks, WeaveNet is easy to train without requiring batch normalization and can be further accelerated by our proposed architecture simplification. Experimental results on PASCAL VOC 2007, PASCAL VOC 2012 benchmarks show signification performance boost brought by WeaveNet. For $320\times320$ input of batch size = 8, WeaveNet reaches 79.5\% mAP on PASCAL VOC 2007 \texttt{test} in 101 fps with only 4 fps extra cost, and further improves to 79.7\% mAP with more iterations.
\end{abstract}

\section{Introduction}

\setcounter{footnote}{=-1}
\begin{figure}
\centering
\includegraphics[width=8cm,height=6cm]{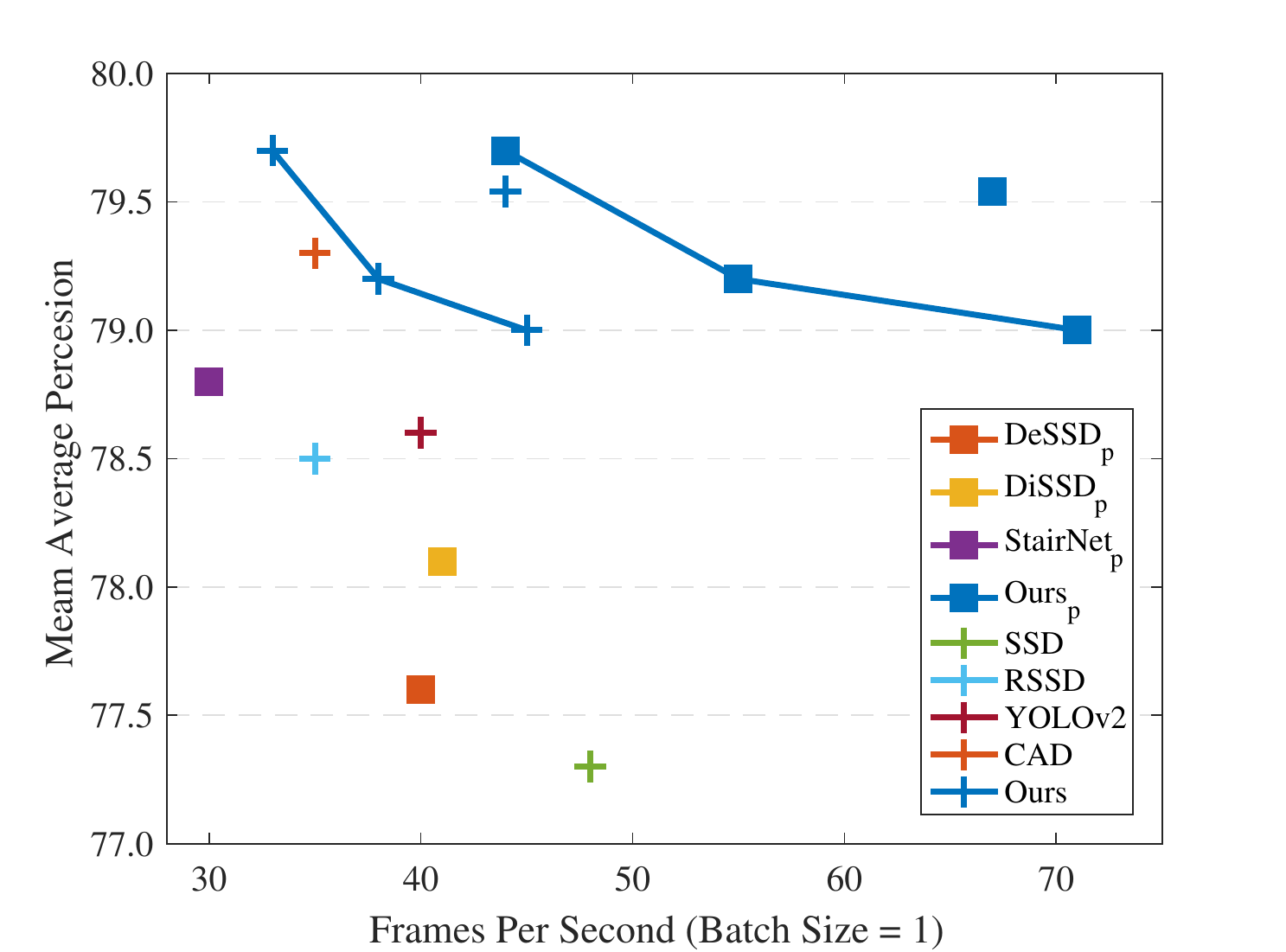}
  \caption[1]{Speed and accuracy trade-off between different models on PASCAL VOC 2007 \texttt{test} dataset. Here, ``$+$'' refers to the performance tested on a single NVIDIA Titan X (Maxwell) GPU, and ``$\blacksquare$'' refers to the performance tested on NVIDIA Tian X (Pascal) GPU, which is a more advanced model with higher computational capability\footnotemark.}
  \label{fig:acc_speed}
  \vspace{-4mm}
\end{figure}

Aggregating multi-scale information is  critical for object detection models to exploit context and achieve better performance in challenging conditions, especially for Single Shot Detector (SSD)~\cite{liu2016ssd}. Different from the conventional two-stage object detectors  (\emph{e.g.}\ RCNN~\cite{girshick2016region}, Faster RCNN~\cite{ren2015faster}) and single-stage detectors (\emph{e.g.}\ YOLO~\cite{redmon2016you} and YOLO-v2~\cite{redmon2016yolo9000}) that detect objects based on features extracted from the last layer, SSD detects objects  from both shallow and deep layers. Under the SSD framework, detectors placed in shallow layers are responsible for detecting small objects, while detectors placed in deeper layers are responsible for detecting larger objects. Such a design significantly improves the detection accuracy on small objects, since fine features from the shallow layers contain richer details in much higher resolution, which may be missed by coarse features from top layers due to down-sampling. However, it still performs not so well on detecting difficult objects, like bottles. This is because the features from shallow layers have a limited receptive filed and all decisions are made based only on the local area. These features cannot perceive global  context information from surrounding areas  for  context reasoning and support making accurate decisions. Integrating information from other scales helps widen the receptive field, thus can alleviate such ambiguities and reduce information uncertainty in the local area.

\footnotetext{CAD~\cite{zhou2017cad} is tested on a GTX 1070 (Pascal) which has similar performance with Nvidia Titan X Maxwell.}

However, the accuracy improvement brought by introducing multi-scale information does not come for free. Existing multi-scale information fusion techniques always introduce  extra components to the existing network which  lead to  significant speed drop. See Figure~\ref{fig:acc_speed}. One of the main extra components consuming a large amount of computational resource is the up-sampling unit which is mostly conducted by deconvolution layers using computational extensive kernels, \emph{e.g.} $4\times4$ kernels. Besides, the extra convolution layers also cost large computational resource, which are used for  gathering and fusing information from different scales. Both of them are essential. The up-sampling unit helps match feature maps to the same scale and the extra convolution layers refine the features before sending them to the detector.  Studies on how to reduce the computation consumption by building a more efficient information fusion unit without degrading performance are still rare.

In this work, we propose to build a light-weight information fusion architecture that can effectively fuse multi-scale information without consuming much computational resource. We reveal that combining information from both lower level finer features and higher level coarser features can lead to more efficient fusion. Then we propose a novel context reasoning architecture which enjoys a smoother information flow by only considering adjacent scales and controllable complexity with an iterative inference mechanism. The proposed light-weight information weaving architecture is called \emph{WeaveNet}, which iteratively conducts multi-scale reasoning with only information from adjacent scales and progressively fuses the long-range information across multiple scales. It does not require batch normalization. Therefore, a deeper backbone network, such as DPN-131~\cite{chen2017dual}, can be adopted for further improving the accuracy as long as the GPU memory can accommodate  1 image per mini-batch. More importantly, WeaveNet is highly efficient and  can gradually improve the performance   by simply performing   more iterations, as shown in Figure~\ref{fig:acc_speed}. We apply WeaveNet for object detection on PASCAL VOC 2007 and 2012 benchmarks. It achieves  79.5\%  mAP on PASCAL VOC 2007 with processing speed as fast as 101 fps.

In the following sections, we will first revisit existing multi-scale methods and highlight the uniqueness of the proposed WeavNet. Then we will introduce the proposed WeaveNet detailedly and evaluate its performance on benchmark datasets.

\section{Related Work}

\FloatBarrier
Recently, the single-stage detector has attracted increasingly more attention due to its simplicity and high detection speed, compared with two-stage detectors, \emph{e.g.} Faster RCNN~\cite{girshick2016region} and RFCN~\cite{dai2016r}. However, despite their advantages, single-stage detectors usually do not perform well for detecting difficult and small objects.

Different from many two-stage detectors and other single-stage models (\emph{e.g.}  YOLO~\cite{redmon2016you}), SSD~\cite{liu2016ssd} detects objects at multiple scales for suiting their sizes: small objects are detected at shallow layers with low-level high-resolution features, while large objects are detected in top layers with high-level low-resolution features. Such a design reduces the demand of using very large input size, \emph{e.g.} $600\times1000$ as commonly used in Faster RCNN~\cite{girshick2016region}, for keeping  rich feature details for the top  layers and thus significantly reduces the computational cost. However, it brings another problem on detecting difficult and small objects. Since features in lower layers have a much smaller receptive field, the network cannot perceive a boarder view to utilize  more global context information and suffer from  ambiguity and insufficient context exploration.

To improve the accuracy of SSD on detecting difficult and small objects, various strategies~\cite{fu2017dssd,lin2017focal,woo2017stairnet,jeong2017enhancement,ren2017accurate} have been proposed to introduce multi-scale information to the conventional SSD framework. One main stream  is to attach a top-down pyramid-like structure to propagate information from top layers to bottom layers to enlarge the receptive filed of each shallow layer. For example, the Deconvolutional Single Shot Detector~\cite{fu2017dssd} uses a deconvolutional module to enlarge the scale of top layers and  adds them back to the  shallow layer features, followed by several extra convolutional layers for fusing the merged information. The very recently proposed StairNet~\cite{woo2017stairnet} and RetinaNet~\cite{lin2017focal} share a similar idea. But they adopt slightly different strategies for choosing a proper adaptive layer before the element-wise sum and they use more effective blocks to conduct further inference on the merged information. However, in order to enable enough information to flow to the final bounding box regressor and  classifier, all of the new attached layers   usually have large input and output channel sizes, \emph{e.g.} $256$, leading to   considerable amount of computational cost.

Another stream of utilizing multi-scale information is to consider both low-level  and high-level information. The main idea is: in addition to introducing   information from top layers to enlarge the receptive fields, they also pass more detailed local information to top layers for making   bound box localization and category inference more precise. However, most of existing methods following this stream   cost more computational resource since information from all scales (other than the target scale) are merged together simultaneously. Rainbow SSD (RSSD)~\cite{jeong2017enhancement} proposes to utilize both low- and high-level information by concatenation, which increases the final fused information from hundreds of channels to 2,816 channels, introducing  significant computational cost in the following layers. Besides, the Recurrent Rolling Convolution Network~\cite{ren2017accurate} proposes to recurrently forward information from top to bottom and bottom to top. However, the inner state-to-state adaptive layers are quite computationally expensive and also significantly slow down the speed compared with the vanilla SSD.

Different from previous works, we propose a weaving structure for multi-scale information fusion. The proposed WeaveNet is naturally friendly to optimization and does not require the batch normalization layer to ease the training. It is also highly parallel in each iteration and costs much lower computation,  which is preferable for real-time application.

\section{WeaveNet}

\subsection{Information Weaving}

\begin{figure*}[t]
\centering
  \includegraphics[width=1.0\textwidth]{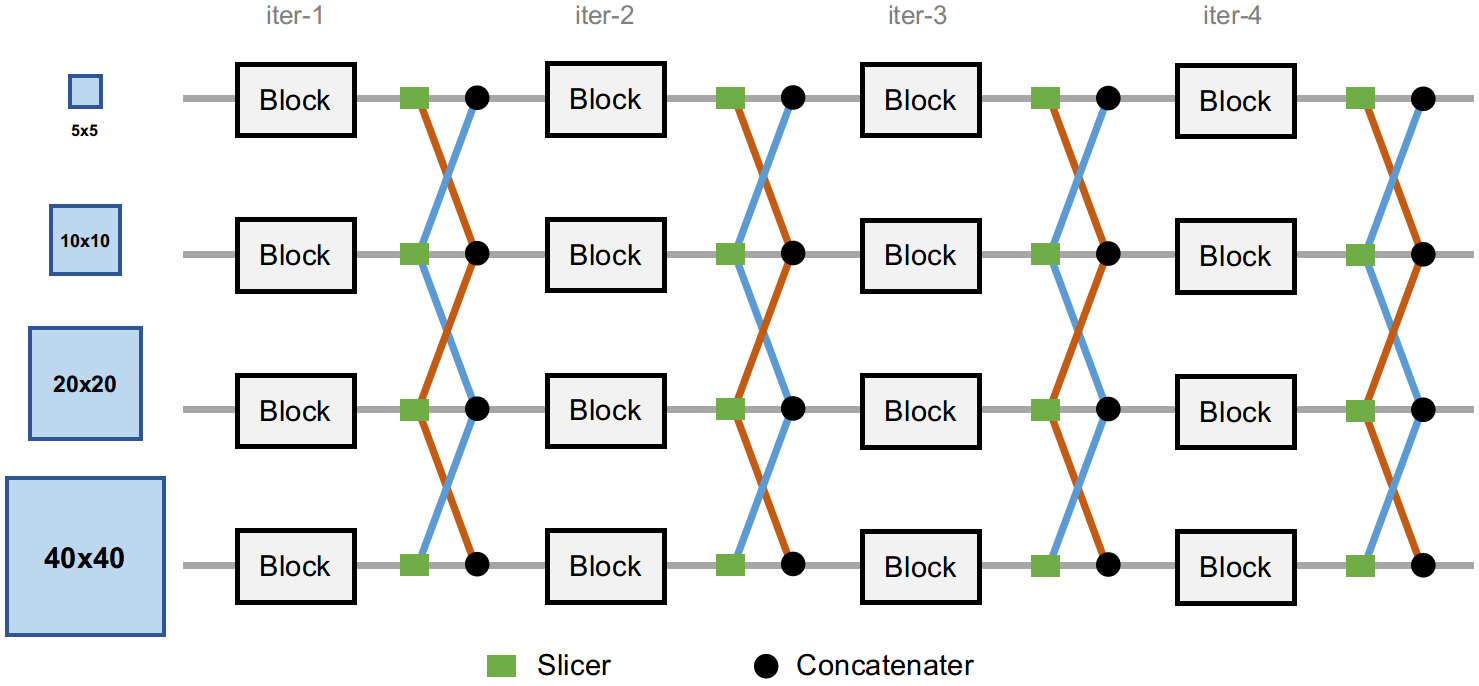}%
  \caption{The overall framework of the proposed WeaveNet. The squares on the left stands for the features extracted from the backbone network, \emph{e.g.} Reduced VGG-16~\cite{liu2015parsenet}; the proposed WeaveNet is constructed by weaving multiple blocks together. The final output is propagated to the final classifier/regressor,  a simple $3\times3$ convolutional layer.}
  \label{fig:weavenet_main}
  \vspace{-4mm}
\end{figure*}

The conventional Single Shot Detector uses features extracted from multiple resolutions to detect objects at various scales. However, as discussed above, the receptive filed in shallower layers can only cover limited local areas, which makes it hard to conduct complex context reasoning based on global features. Moreover, features from higher layers passing through several down-sampling stages may lose detailed information for precise localization. In this work, we propose to introduce features from both lower and higher layers through a novel information weaving architecture to overcome both drawbacks without efficiency decline.

As shown in Figure \ref{fig:weavenet_main}, the idea of the proposed information weaving structure is to gradually weave the information from adjacent scales for the detector in the current scale. Here, the ``gradually'' means that only  information from adjacent scales is taken into account, since we believe current scale should focus more on adjacent scales instead of those faraway at the very first iterations.  We propose to integrate  the long-range information by an iterative inference process, allowing the information  to propagate from neighbors to its neighbors  to the  current scale. By weaving information iteratively, sufficient multi-scale context  information can  be transferred and integrated to the current scale thoroughly.

\subsection{Network Architecture}

\begin{figure}
\centering
  \includegraphics[width=0.45\textwidth]{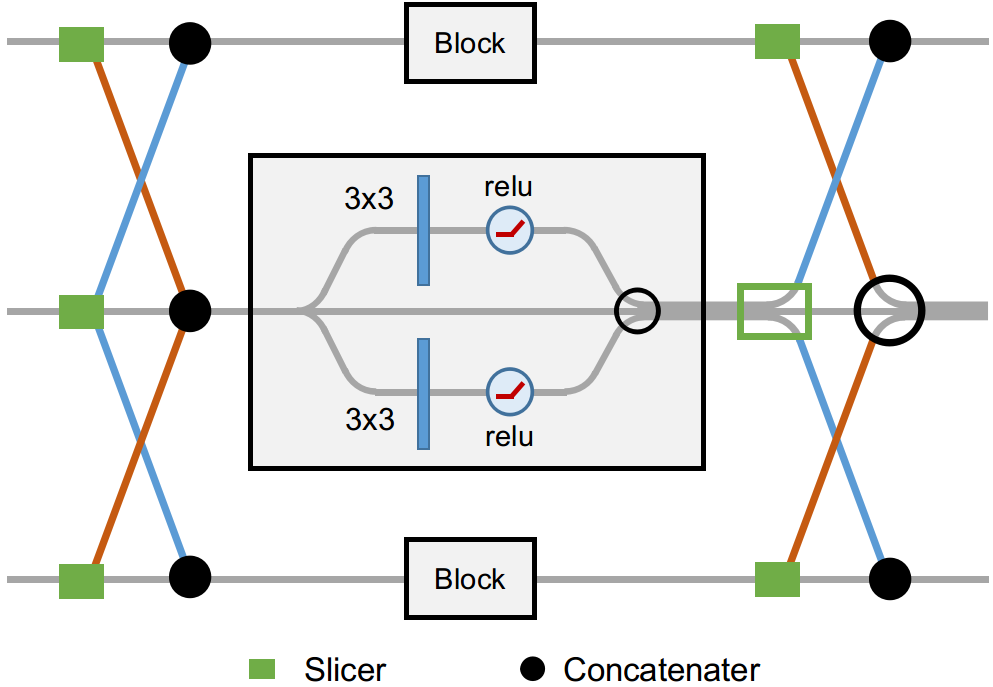}%
  \caption{The inner architecture of each WeaveNet block. It contains two $3\times3$ convolutional layers and each of them is followed by a ReLu activation layer. The transformed features are then sent to the upper and lower scales separately.}
  \label{fig:zoom_in}
\end{figure}

The overall architecture of the proposed WeaveNet is shown in Figure~\ref{fig:weavenet_main} and we disclose the inner structure of each block in Figure~\ref{fig:zoom_in}.

In the main architecture, the right part shows our proposed context information weaving structure, while the  left part shows the raw features extracted from the backbone CNN. The WeaveNet takes the raw features extracted from the backbone network as input and outputs the refined features. Each of the refined features is then attached by two separate convolution layers for bounding box regression and classification respectively, same as the vanilla SSD. Within the proposed WeaveNet, information are gathered and fused iteratively. Each scale extracts useful information from both of its higher adjacent layers (with low resolution) and lower  adjacent layer (with high resolution). These information is  added to the current state for next decision making after going through a \textit{block} for information fusion and inference. During the iterative refinement, besides propagating information to longer range scales, the information can also propagate back to its own scale for making more complex reasoning and introducing greater non-linearity. To make information from adjacent scales be compatible with current scale, we use a bi-linear interpolation for enlarging the feature maps and use $2\times2$ max pooling with stride $= 2$ for reducing the feature maps. Both of them are parameter-free and introduce negligible computational cost.

Details about the inner block architecture are shown in Figure \ref{fig:zoom_in}, which is designed to be as light as possible for reducing computational cost. At the beginning of the bock, information from both higher and lower layers are concatenated   together to form  inputs to the block, which contains information from all previous iterations including the raw features from the backbone network. Then, all the information is passed though a $3\times3$ convolution layer with a ReLU activation layer for a spatial non-linear fusion. The output is then passed to its neighborhood for next iteration. We    repetitively stack these simple blocks for modeling and aggregating more complex and richer context information. Based on our experiments, the most significant improvement comes from the 1st iteration and usually its maximum performance is achieved at the 5th iteration.

\paragraph{Model details}
We use a reduced VGG16 network as the backbone and attach the proposed WeaveNet for multi-scale information fusion. The overall setting follows the original SDD for fair comparison, where we extract the features from \texttt{conv4\_3}, \texttt{fc6}, and add 6 more layers after the \texttt{fc6}. In our training platform, the existing bi-linear interpolation is implemented by using channel-wise devconvolution, which can only support integer scale factor as the pooling layer. Thus, we set the input image size as $320\times320$, so that the size of each scale is $\{40, 20, 10, 5, 3, 1\}$ respectively   where  the scales are reduced by a factor of 2 in the first 4 scales. In this way,    it is easier to attach our proposed WeaveNet. Since the last two scales are  small, one $3\times 3$ convolution kernel is able to  cover the whole map. Hence, we do not refine them further and keep them the same as the vanilla SSD.

\subsection{Architecture Simplification}

Single Shot Detector is known for its fast speed as well as high accuracy. Losing either of these advantages would make the detector less favorable. In this subsection, we elaborate on how to accelerate the WeaveNet by grouping fragmented computation together to reduce the data allocation and communication cost and increase the hardware usage rate by reformulating the network topology.

Suppose $f_{i}^{0}$ is the raw feature extracted from the $i$-th scale of the backbone network. Let $f_{i}^{t} = [\hat{f}_{i-1}^{t-1}; f_{i}^{t-1}; \check{f}_{i+1}^{t-1}]$ be the input and $[\check{f}_{i}^{t},f_{i}^{t}, \hat{f}_{i}^{t}]$ be outputs of the \textit{Block} shown in Figure~\ref{fig:zoom_in} in the $t$-th iteration. Here $\check{f}_{i}^{t}$ will be sent to the lower scale and $\hat{f}_{i}^{t}$ will be sent to the upper scale. Then, the convolution operation in Figure \ref{fig:zoom_in} can be formulated as
\begin{equation}
\label{eq:1}
\begin{aligned}
\check{f}_{i}^{t} & = \sigma( [\hat{f}_{i-1}^{t-1}; f_{i}^{t-1}; \check{f}_{i+1}^{t-1}] * \check{W}^{t} + \check{b} ), \\
\hat{f}_{i}^{t} & = \sigma( [\hat{f}_{i-1}^{t-1}; f_{i}^{t-1}; \check{f}_{i+1}^{t-1}] * \hat{W}^{t} + \hat{b} ), \\
\end{aligned}
\end{equation}
where $\check{W}$, $\check{b}$ are the convolutional kernel and bias of the lower $3\times3$ layer respectively, $\hat{W}$, $\hat{b}$ are the convolutional kernel and bias for the upper $3\times3$ layer, and $\sigma(\cdot)$ is the ReLu activation function shown in Figure \ref{fig:zoom_in}.

We propose to  group the computation by $W^{t} = [\check{W}^{t}; \hat{W}^{t}]$ and $b^{t} = [\check{b}^{t}; \hat{b}^{t}]$. Then,  Eqn.~\eqref{eq:1} can be simplified as
\begin{align*}
[\check{f}_{i}^{t},\hat{f}_{i}^{t}] = \sigma([\hat{f}_{i-1}^{t-1}; f_{i}^{t-1}; \check{f}_{i+1}^{t-1}] * W^{t} + b).
\end{align*}
Further splitting   $W^{t}$ into $W^{t}=[W_a^{t}, W_b^{t}]$ gives
\begin{align*}
&[\check{f}_{i}^{t},\hat{f}_{i}^{t}] \\
 = & \sigma([\hat{f}_{i-1}^{t-1}; f_{i}^{t-1}; \check{f}_{i+1}^{t-1}] * W^{t} + b)  \\
								  = &\sigma([\hat{f}_{i-1}^{t-1};...;\hat{f}_{i-1}^{1}; f_{i}^{0};\check{f}_{i+1}^{1};...;\check{f}_{i+1}^{t-1}] * W^{t} + b) \\
								  = & \sigma([\hat{f}_{i-1}^{t-1};...;\hat{f}_{i-1}^{1};\check{f}_{i+1}^{1};...;\check{f}_{i+1}^{t-1}] * W_a^{t} + b + f_{i}^{0} * W_b^{t}).
\end{align*}
The computation in each block can be separated into two parts. One depends on the previous states and the other part does not. Thus the latter part can be grouped and pre-computed for further acceleration, \emph{i.e.} $[f_{i}^{0} * W_b^{1}, ..., f_{i}^{0} * W_b^{t}] = f_{i}^{0} * W_{b}$.

\begin{figure}
\centering
  \includegraphics[width=0.45\textwidth]{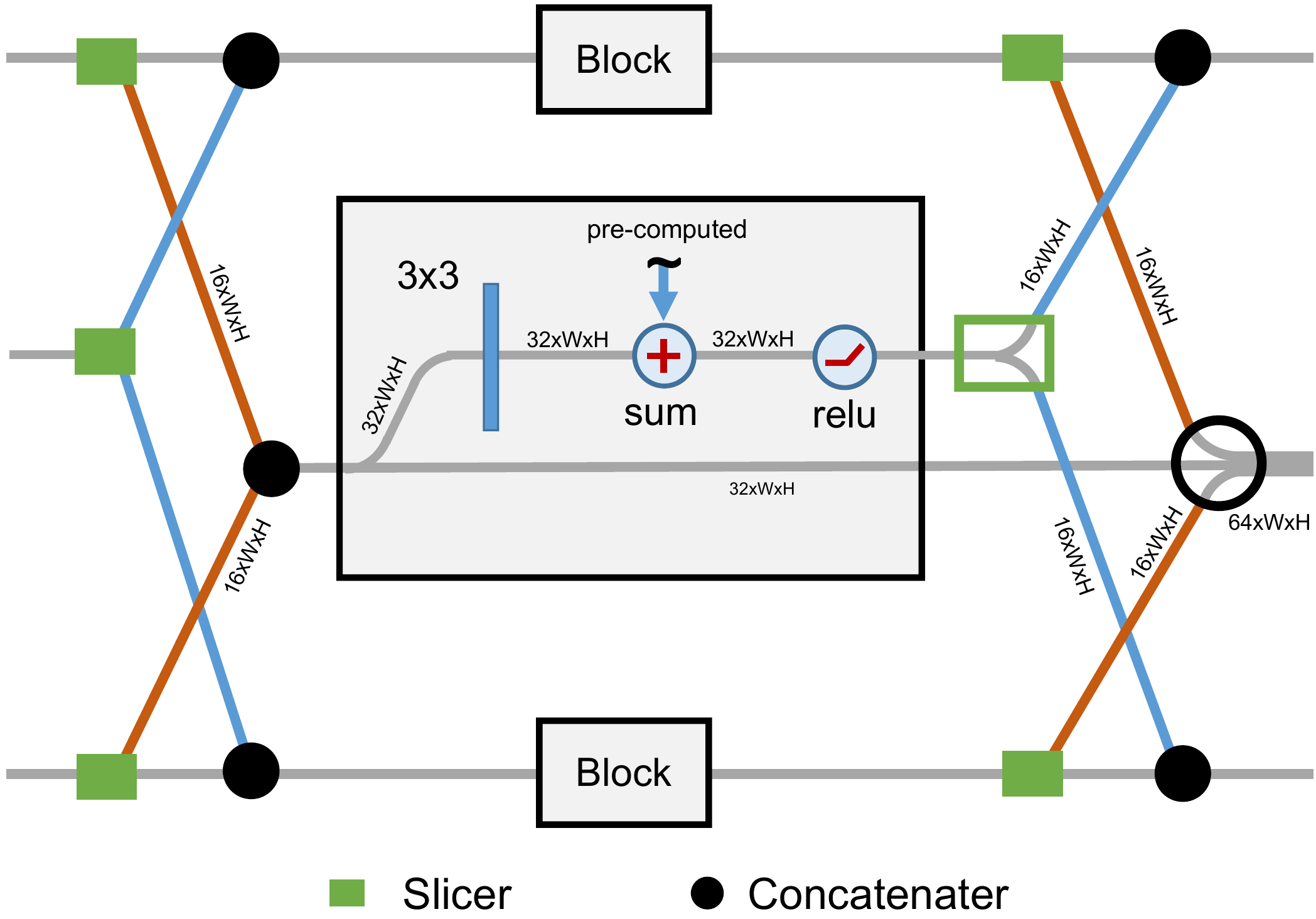}
  \caption{Simplified inner architecture of each WeaveNet block. ``$\sim$'' refers to the pre-computed result which does not depend on current input features and ``$\oplus$'' stands for the element-wise sum.}
  \label{fig:simplified}
\end{figure}

Figure \ref{fig:simplified} shows the corresponding simplified architecture, where fragmented computations are grouped together to reduce the unnecessary data allocation and communication cost and increase the hardware usage rate for acceleration. More specifically, different from the topology shown in Figure \ref{fig:zoom_in},  only the outputs from adjacent scales are gathered by concatenation. Within the block, these inputs would all be sent to a single transformer. The output is  element-wisely summed with a pre-computed source and then split and sent to its adjacent scales. Both inputs and outputs of each block are usually in low dimension,  \eg~{32}.  When only considering the pure computational cost, such computational cost of adding one block can be more than $ 10\times  $  less than adding a single $3\times3$ layer with both input and output channels equal to $256$ for fusing multi-scale information.

\section{Experiments}

We implement the proposed WeaveNet based on vanilla SDD~\cite{liu2016ssd} using the same version of Caffe~\cite{jia2014caffe}. Following the original SSD, we adopt the same optimization strategy and train the same number of iterations for fair comparison. In particular, all networks are trained with batch size of 32, and the learning rate is reduced from $4\times10^{-3}$ to $4\times10^{-5}$ by $10^{-1}$ at the $80$K and $100$K iterations and terminated at the $120$K iterations. Frames Per Second (fps) is reported on both Nvidia Titan X (Maxwell) GPU and Nvidia Titan X (Pascal) GPU with the same NVIDIA Library, \emph{i.e.} CUDA 8.0 $+$ cuDNN v5.1.

In the experiments, we evaluate the WeaveNet on the widely used PASCAL VOC 2007 and PASCAL VOC 2012 benchmarks~\cite{everingham2010pascal}. In order to provide more insights, we first conduct controlled experiments on the PASCAL VOC 2007 benchmarks to study the properties of WeaveNet. Then, based on the results, we test the proposed WeaveNet with its best settings on both benchmarks and compare it with existing state-of-the-art object detection methods through an in-depth analysis.

\subsection{Importance of Coarse and Fine Features}

We start with an ablation study to investigate the importance of top-down  and bottom-up information propagation. Since   top-down information propagation has already been observed to be important by many papers~\cite{jeong2017enhancement,lin2016feature,lin2017focal,woo2017stairnet}, in our experiments, we are more concerned about whether further introducing finer features from lower layers would improve the accuracy.

\begin{table*}
\renewcommand{\arraystretch}{1.4}
\centering
	\caption{Ablation study on the importance of using the coarse and fine features. The ``Top-down'' experiment studies the importance of introducing high-level coarse features from top layers, and the ``Bottom-up'' studies the importance of introducing low-level fine features from bottom layers. Performance  is measured by mean of Average Precision (mAP) on PASCAL VOC 2007 test set.}
	\label{table:ablation_study:feature}	
	\vspace{1mm}
	{
	\footnotesize
	\begin{tabular}{c|c|c|c|c|c|c|c|c}
		\hline
		Method    & Settings	 	& Input Size	 	 & Top-down 	  & Bottom-up  	& mAP (small) & mAP (medium) & mAP (large) & mAP (overall)
		\\ \hline
		SSD~\cite{liu2016ssd}	  &  N/A    	    & $320\times320$ & 			  & 				& 43.4\% 	  & 74.1\% 		& 79.2\%  	  & 77.3\%
		\\
		WeaveNet  & k=16, iter=1 & $320\times320$ & \checkmark &   			& 47.2\% 	  & 75.4\% 		& 79.7\%  	  & 78.6\%
		\\
		WeaveNet  & k=16, iter=1 & $320\times320$ &  		  & 	\checkmark 	& 43.7\% 	  & 75.0\% 		& 79.7\%  	  & 78.1\%
		\\
		WeaveNet  & k=16, iter=1 & $320\times320$ & \checkmark & 	\checkmark 	& 47.4\% 	  & 76.3\% 		& 78.6\%  	  & 79.0\%
		\\ \hline
	\end{tabular}
	}
\end{table*}

The ablation study is designed by blocking either the bottom-up information path or the top-down information path individually to study effects of different components. In the experiments,   the number of iteration is set to be $1$ for WeaveNet. Table~\ref{table:ablation_study:feature} shows the experiment results. As can be seen from the last column, the ``top-down'' information propagation improves the overall mAP by $1.3\%$ as expected, while the ``bottom-up'' information propagation can also improve the overall mAP by $0.8\%$. By utilizing both top-down and bottom-up information, it further improves the mAP to $79.0\%$, indicating that both top-down and bottom-up feature are important.

To further investigate effects of introducing top and bottom features, we follow \cite{woo2017stairnet} to sort the testing images into different scales by the area of ground truth bounding box and evaluate mAP w.r.t.\ specific object size. Specifically, the ground truth bounding boxes are divided into three parts per class: \ie $\text{small:} [0 \sim 25\%), \text{medium:} [25\% \sim 75\%), \text{large:} [75\%\sim 100\%]$. When doing evaluation on a specific scale, ground truth bounding box on other scales are ignored. The results are shown in the 6th to 8th columns in Table~\ref{table:ablation_study:feature}. As can be seen from the results, the ``Top-down'' information significantly benefits small object detection. Further introducing the fine detailed features from bottom layers help detect medium objects. However, it is interesting to see that once both top and bottom features are considered, the detection accuracy on large objects  slightly drops.

\begin{figure}
\centering
\resizebox{0.45\textwidth}{!}{
  \includegraphics{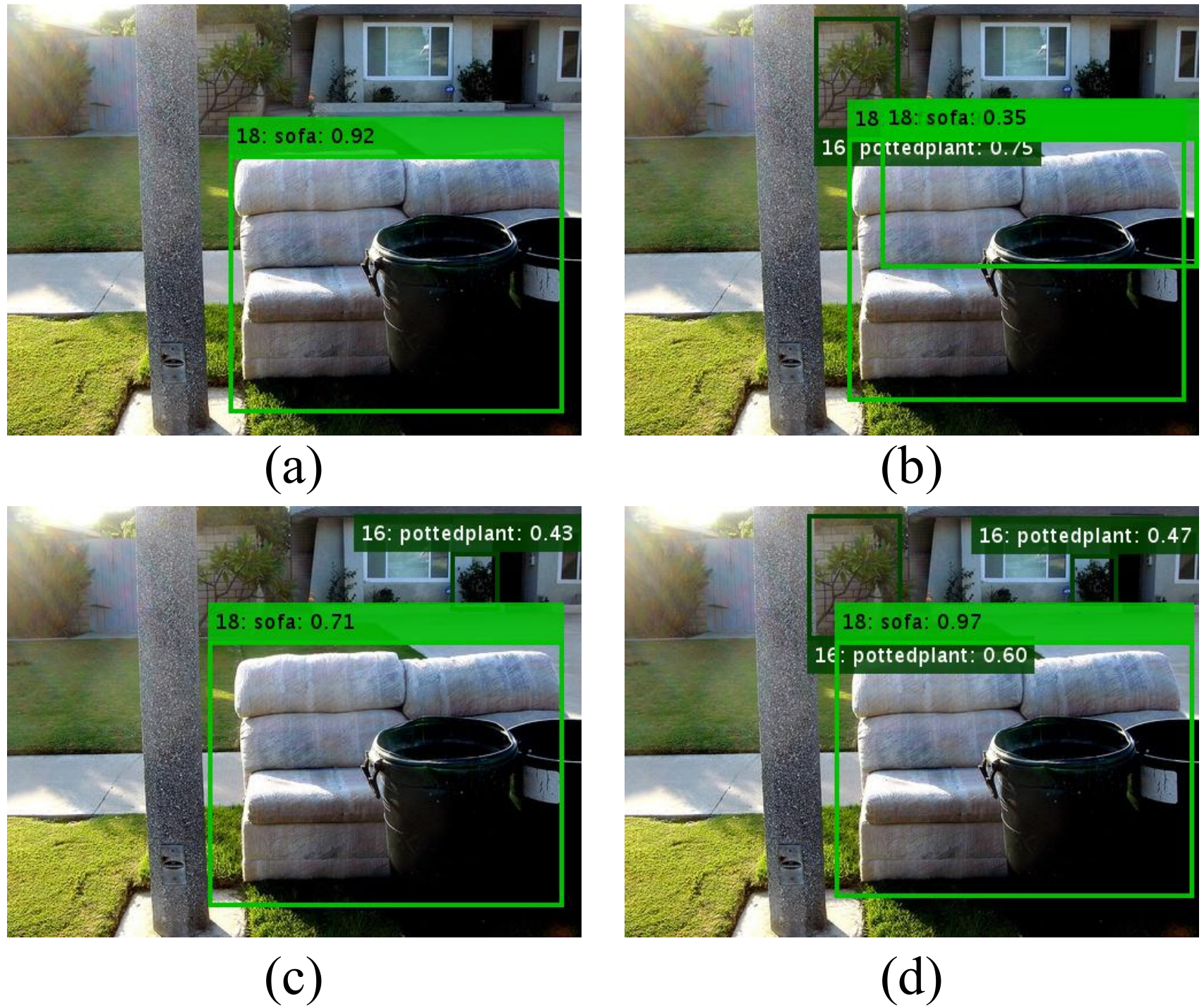}%
  }
  \caption{Comparisons on detection quality among four different fusion strategies. (a)  Vanilla SSD (w/o fusion); (b)  Top-down fusion; (c)  Bottom-up fusion; (d)  Weaving both top-down and bottom-up information.}
  \label{fig:fine_coarse_fea}
\end{figure}

We visualize the testing results in Figure~\ref{fig:fine_coarse_fea} to compare the detection results of each detector. As can be seen from Figure~\ref{fig:fine_coarse_fea}~(d), smaller and more difficult objects are detected correctly, compared with other strategies in Figure~\ref{fig:fine_coarse_fea} (a)-(c).

The above ablation studies verify our conjecture that combining both low-level feature and high-level features can further improve object detection accuracy. Thus, in the following experiments, we use the full version of WeaveNet to further study its properties and compare it with state-of-the-arts.

\subsection{Effectiveness and Efficiency}

The most attractive advantage of using a single stage detector is the speed. Different from the two-stage object detectors which require an object proposal generation stage and an object prediction stage, the single stage detector directly performs the prediction and thus saves a huge amount of computation resource. It would be less useful if the attached multi-scale fusion component loses the speed advantage with only slight performance improvement. In experiments here, we first study the performance improvement brought from each individual component and then adjust the number of iteration in the WeaveNet to study the speed-accuracy tread-off comparing with state-of-the-art multi-scale fusion methods.

\begin{table}
\renewcommand{\arraystretch}{1.4}
\centering
\caption{Ablation study of WeavNet on PASCAL VOC 2007 test set. Speed (in fps) is tested on a single NVIDIA Titan X (Maxwell) with batch size equal to 1. The performance is measured by mean of Average Precision (mAP).}
\label{table:ablation_study:component}	
\vspace{1mm}
\resizebox{0.48\textwidth}{!}
{
	\begin{tabular}{c|cc|cc|ccc|c}
	\hline
	Component
		& \multicolumn{7}{c|}{WeaveNet} & SSD  \\ \hline
	bbox refinement
		& \checkmark	& \checkmark	&    		&    		&   			&   			&   			&      \\
	use anchor {[}2,3{]}
		& \checkmark& \checkmark	& \checkmark	& \checkmark	&   			&   			&   			&      \\ \hline
	iter=1, k=64
		& \checkmark	& 			& \checkmark	&			& \checkmark	&   			&   			&      \\
	iter=1, k=32
		&    		&    		&   			&   			&   			& \checkmark	&   			&      \\
	iter=1, k=16
		&    		& \checkmark	&    		& \checkmark	&   			&   			& \checkmark	&      \\ \hline
	\begin{tabular}[c]{@{}c@{}}fps\end{tabular}
		&    42	    &   	  45		&    42   	&   	 45		&    43  	&    44  	&     46	    &  48  \\ \hline
	mAP (\%)
		&  79.25		&    78.96	&    79.03	&   	78.95	&   	79.02	&   78.69	&   	78.60	& 77.33 \\
	\hline
	\end{tabular}
}
\end{table}

\paragraph{Width of WeaveNet} To study the effectiveness of introducing different amount of scale information from adjacent scales, we vary the channels size of concatenated adjacent channels, noted as $k$, from $k=16$ to $k=64$. The larger the number is, the richer information can be introduced for each iteration. As can be seen from the fourth column Table~\ref{table:ablation_study:component}, the accuracy consistently improves. However, the speed is slightly decreased because of the higher computational complexity.

\paragraph{Number of anchor boxes} The regressor of single shot detector predicts the offset w.r.t.\ its default anchors. The original anchor setting is to use 5 different aspect ratios, \emph{i.e.} $[1/2, 1, 2]$, for the first one and last two scales, and use 3  aspect ratios \emph{i.e.} $[1/3, 1/2, 1, 2, 3]$, for the middle scales. However, as can be seen from the third column of Table~\ref{table:ablation_study:component}, introducing more anchors by using five kinds of aspect ratios for all six scales can also improve the accuracy by about $0.4\%$, with a slight computation overhead.

\paragraph{Bounding box refinement} We refine the final bounding box location by conducting a bounding box refinement after the NMS stage. Instead of directly using the NMS output, we further refine the location of each bounding box using a weighted sum with its surrounding boxes whose IoU is greater than 0.6. The weight is set to be the score of each box. We deploy the bounding box refinement upon a well trained model. The 2nd column in Table~\ref{table:ablation_study:component} shows improvement of the proposed bounding box refinement technique. In our experiment, we find the proposed refinement technique can always help gain $0.1\%-0.3\%$ mAP compared with bounding box without refinement.

\begin{table}
\renewcommand{\arraystretch}{1.4}
\centering
	\caption{Speed accuracy tread-off of different methods on PASCAL VOC 2007 test set. Results with two batch sizes, \ie,   1 and 8,  are reported. The performance is measured by mean of Average Precision (mAP, in \%).}
	\label{table:iter_inference}	
	\vspace{1mm}	
	\resizebox{0.48\textwidth}{!}
	{
\begin{tabular}{ccccccc}
\hline
\multicolumn{1}{c|}{\multirow{2}{*}{Method}}                                       & \multicolumn{1}{c|}{\multirow{2}{*}{Backbone}} & \multicolumn{2}{c|}{fps (Maxwell)}                            & \multicolumn{2}{c|}{fps (Pascal)}                             & \multicolumn{1}{c}{\multirow{2}{*}{mAP (\%)}} \\ \cline{3-6}
\multicolumn{1}{c|}{}                                                              & \multicolumn{1}{c|}{}                          & \multicolumn{1}{c|}{Size = 1} & \multicolumn{1}{c|}{Size = 8} & \multicolumn{1}{c|}{Size = 1} & \multicolumn{1}{c|}{Size = 8} & \multicolumn{1}{c}{}
\\ \hline
\multicolumn{1}{c|}{SSD~\cite{liu2016ssd}}                                                           & \multicolumn{1}{c|}{VGG16}                    & \multicolumn{1}{c|}{48}       & \multicolumn{1}{c|}{67}       & \multicolumn{1}{c|}{75}       & \multicolumn{1}{c|}{105}      & \multicolumn{1}{c}{77.3}
\\
\multicolumn{1}{c|}{DiSSD~\cite{xiang2017context}}                                                         & \multicolumn{1}{c|}{VGG16}                    & \multicolumn{1}{c|}{--}       & \multicolumn{1}{c|}{--}       & \multicolumn{1}{c|}{41}       & \multicolumn{1}{c|}{--}       & \multicolumn{1}{c}{78.1}
\\
\multicolumn{1}{c|}{RSSD~\cite{jeong2017enhancement}}                                                          & \multicolumn{1}{c|}{VGG16}                    & \multicolumn{1}{c|}{35}       & \multicolumn{1}{c|}{--}       & \multicolumn{1}{c|}{--}       & \multicolumn{1}{c|}{--}       & \multicolumn{1}{c}{78.5}
\\
\multicolumn{1}{c|}{DSSD~\cite{fu2017dssd}}                                                          & \multicolumn{1}{c|}{ResNet101}                & \multicolumn{1}{c|}{10}       & \multicolumn{1}{c|}{--}       & \multicolumn{1}{c|}{--}       & \multicolumn{1}{c|}{--}       & \multicolumn{1}{c}{78.6}
\\
\multicolumn{1}{c|}{StairNet~\cite{woo2017stairnet}}                                                      & \multicolumn{1}{c|}{VGG16}                    & \multicolumn{1}{c|}{--}       & \multicolumn{1}{c|}{--}       & \multicolumn{1}{c|}{30}       & \multicolumn{1}{c|}{--}       & \multicolumn{1}{c}{78.8}
\\ \hline
\multicolumn{1}{c|}{\begin{tabular}[c]{@{}c@{}}WeaveNet\\ (k=16, iter=1)\end{tabular}} & \multicolumn{1}{c|}{VGG16}                    & \multicolumn{1}{c|}{45}       & \multicolumn{1}{c|}{63}       & \multicolumn{1}{c|}{71}       & \multicolumn{1}{c|}{104}      & \multicolumn{1}{c}{79.0}
\\
\multicolumn{1}{c|}{\begin{tabular}[c]{@{}c@{}}WeaveNet\\ (k=32, iter=1)\end{tabular}} & \multicolumn{1}{c|}{VGG16}                    & \multicolumn{1}{c|}{44}       & \multicolumn{1}{c|}{61}       & \multicolumn{1}{c|}{67}       & \multicolumn{1}{c|}{101}      & \multicolumn{1}{c}{79.5}
\\
\multicolumn{1}{c|}{\begin{tabular}[c]{@{}c@{}}WeaveNet\\ (k=16, iter=3)\end{tabular}} & \multicolumn{1}{c|}{VGG16}                    & \multicolumn{1}{c|}{38}       & \multicolumn{1}{c|}{56}       & \multicolumn{1}{c|}{55}       & \multicolumn{1}{c|}{82}       & \multicolumn{1}{c}{79.2}
\\
\multicolumn{1}{c|}{\begin{tabular}[c]{@{}c@{}}WeaveNet\\ (k=16, iter=5)\end{tabular}} & \multicolumn{1}{c|}{VGG16}                    & \multicolumn{1}{c|}{33}       & \multicolumn{1}{c|}{50}       & \multicolumn{1}{c|}{44}       & \multicolumn{1}{c|}{72}       & \multicolumn{1}{c}{79.7}
\\ \hline
\end{tabular}
	}
\end{table}

\paragraph{Iterative information weaving} One major attractive property of the proposed information weave structure is: by gradually weaving information from different scales, the receptive filed is enlarged and the learning ability of the whole network is increased. To investigate effectiveness of such an iterative inference procedure, we design another set of experiments by varying the number of iterations of the proposed WeaveNet while keeping other components fixed. The results are summarized in Table~\ref{table:iter_inference}. As can be seen from the results, the accuracy keeps increasing along with  more iterations, demonstrating the proposed information weaving architecture can effectively fuse and refine the multi-scale information. The performance boost is also observed for other settings consistently. We also plot the results in Figure~\ref{fig:acc_speed}, where the solid line represents the same WeaveNet but using different number of iterations. The speed does gradually decrease when using more iterations. However, when comparing with other existing multi-scale fusion techniques  in the first block of Table~\ref{table:iter_inference},  WeaveNet still shows significant superiority in both speed and accuracy.

\subsection{Results on PASCAL 2007}

In this subsection, we show the performance comparison between the WeaveNet and  state-of-the-art object detection models. All of the models are trained on the union set of PASCAL VOC 2007 trainval and VOC 2012 trainval, and evaluated on PASCAL VOC 2007 \texttt{test} set.

\renewcommand{\arraystretch}{1.5}
\begin{table*}[t]
	\centering
	\caption{Object detection results on PASCAL VOC 2007 \texttt{test} set. The performance is measured by mean of Average Precision (mAP, in \%).	 }
	\label{table:VOC07}
	\resizebox{\textwidth}{!}{
    \setlength\tabcolsep{2.5pt}
	\begin{tabular}{c|c|c|cccccccccccccccccccc}
		\toprule 
		\textbf{Two-Stage} & Backbone
		& \textbf{mAP}
		& areo   & bike   & bird   & boat   & bottle & bus    & car    & cat    & chair  & cow
		& table  & dog    & horse  & mbk    & prsn   & plant  & sheep  & sofa   & train  & tv   \\
		\midrule 
		Faster~\cite{ren2015faster} & VGG16
		& 73.2
		& 76.5   & 79.0   & 70.9   & 65.5   & 52.1   & 83.1   & 84.7   & 86.4   & 52.0   & 81.9
		& 65.7   & 84.8   & 84.6   & 77.5   & 76.7   & 38.8   & 73.6   & 73.9   & 83.0   & 72.6 \\
		ION~\cite{bell2016inside} & VGG16
		& 75.6
		& 79.2   & 83.1   & 77.6   & 65.6   & 54.9   & 85.4   & 85.1   & 87.0   & 54.4   & 80.6
		& 73.8   & 85.3   & 82.2   & 82.2   & 74.4   & 47.1   & 75.8   & 72.7   & 84.2   & 80.4 \\
		MR-CNN~\cite{gidaris2015object} & VGG16
		& 78.2
		& 80.3   & 84.1   & 78.5   & 70.8   & 68.5   & 88.0   & 85.9   & 87.8   & 60.3   & 85.2
		& 73.7   & 87.2   & 86.5   & 85.0   & 76.4   & 48.5   & 76.3   & 75.5   & 85.0   & 81.0 \\
		Faster~\cite{he2016deep} & ResNet101
		& 76.4
		& 79.8   & 80.7   & 76.2   & 68.3   & 55.9   & 85.1   & 85.3   & 89.8   & 56.7   & 87.8
		& 69.4   & 88.3   & 88.9   & 80.9   & 78.4   & 41.7   & 78.6   & 79.8   & 85.3   & 72.0 \\
		R-FCN~\cite{dai2016r} & ResNet101
		& 80.5
		& 79.9   & 87.2   & 81.5   & 72.0   & 69.8   & 86.8   & 88.5   & 89.8   & 67.0   & 88.1
		& 74.5   & 89.8   & 90.6   & 79.9   & 81.2   & 53.7   & 81.8   & 81.5   & 85.9   & 79.9 \\
		\bottomrule
		\toprule 
		\textbf{One-Stage} & Backbone
		& \textbf{mAP}
		& areo   & bike   & bird   & boat   & bottle & bus    & car    & cat    & chair  & cow
		& table  & dog    & horse  & mbk    & prsn   & plant  & sheep  & sofa   & train  & tv   \\
		\midrule
		SSD300*~\cite{liu2016ssd} & VGG16
		& 77.5
		& 79.5   & 83.9   & 76.0   & 69.6   & 50.5   & 87.0   & 85.7   & 88.1   & 60.3   & 81.5
		& 77.0   & 86.1   & 87.5   & 83.9   & 79.4   & 52.3   & 77.9   & 79.5   & 87.6   & 76.8 \\
		DSSD 321~\cite{fu2017dssd} & ResNet101
		& 78.6
		& 81.9   & 84.9   &\bf{80.5}& 68.4  & 53.9   & 85.6   & 86.2   &\bf{88.9}& 61.1   & 83.5
		& 78.7   & 86.7   &\bf{88.7}&\bf{86.7 }& 79.7   & 51.7   & 78.0  &\bf{80.9}& 87.2   &\bf{79.4}\\
		I-SSD~\cite{ning2017inception} & VGG16
		& 78.6
		& 82.4   & 84.3   & 78.1   & 70.6   & 52.8   & 85.7   & 86.8   & 88.3    & 62.4    & 82.7
		& 78.0   & 86.7   & 88.3   & 86.0   & 79.9   & 53.4   & 78.5   &\bf{80.9}&\bf{88.5}& 77.8 \\	
		StairNet~\cite{woo2017stairnet} & VGG16
		& 78.8
		& 81.3   & 85.4   & 77.8   & 72.1   &\bf{59.2}& 86.4  & 86.8   & 87.5   &\bf{62.7}&\bf{85.7}
		& 76.0   & 84.1   & 88.4   & 86.1   & 78.8   & 54.8   & 77.4   & 79.0   & 88.3    & 79.2 \\	
		\bf{WeaveNet} & VGG16
		& \bf{79.7}
		&\bf{83.0}&\bf{88.1}& 79.6 &\bf{72.7}& 57.7    &\bf{87.1}&\bf{86.9}& 88.7 & 62.5   & 84.6
		&\bf{79.0}&\bf{87.1}& 86.9 & 85.1    &\bf{80.4}&\bf{56.2}&\bf{82.4}& 80.2 & 87.9   & 78.8 \\
		\bottomrule 
	\end{tabular}
    }
\end{table*}
\renewcommand{\arraystretch}{1.2}


Table \ref{table:VOC07} shows the results of comparing WeaveNets with the state-of-the-art models. As can be seen from the results, WeaveNet achieves $79.7\%$ mAP, much higher than the $77.5\%$ mAP achieved by vanilla SSD. Comparing with other state-of-the-art multi-scale fusion methods, \emph{e.g.} StariNet, WeaveNet further improves the mAP by $0.9\%$. We note that   WeaveNet   in Table~\ref{table:VOC07} has $k = 16$ and $iter = 5$, which is slower than WeaveNets using less  iterations. As can be seen in Figure~\ref{fig:acc_speed}, we found the   WeaveNet with $k = 32$ and $iter = 1$ actually achieves slightly better speed and accuracy tread-off, with $mAP = 79.5$ in $67$ fps (batch size = 1) using TITAN X (PASCAL) GPU.

\subsection{Results on PASCAL 2012}

\begin{figure}[t!]
	\center
	\resizebox{0.5\textwidth}{!}{
		\includegraphics{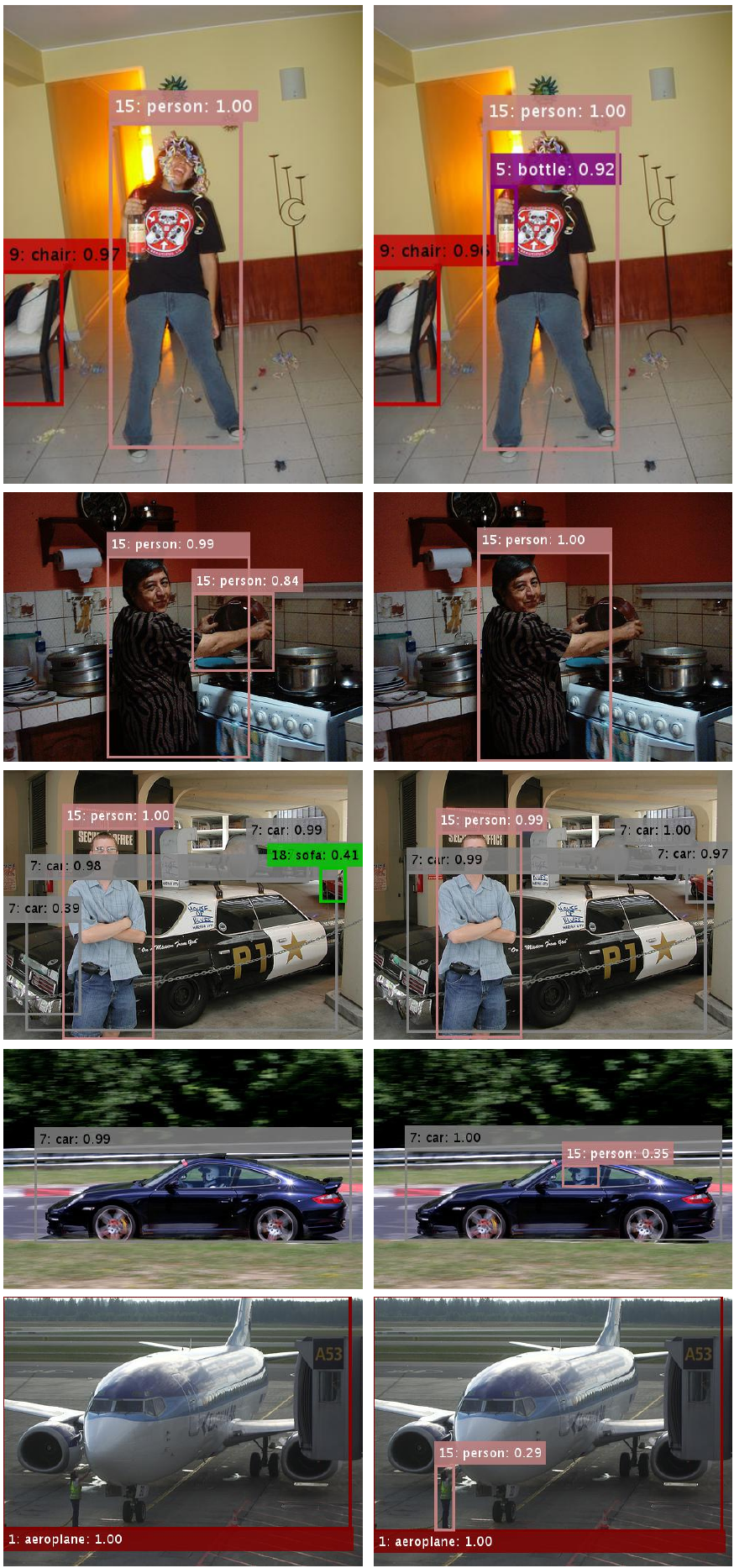}
	}	
	\caption{Detection result comparison between   the vanilla SSD (left) and the proposed WeaveNet(right) on VOC 2012 \texttt{test} set.}
	\label{fig:VOC12_viz}
\end{figure}

We also evaluate the proposed method on PASCAL VOC 2012 benchmark, where more training samples are included and the testing set is replaced by a more difficult set which has about 2 times more testing images than PASCAL VOC 2007. We did not extensively tune the training parameter and all models are trained in the same number of iterations following exactly the same learning rate and weight decay as used on PASCAL VOC 2007. The training set is a union of PASCAL VOC 07 trainval + test and PASCAL VOC 2012 trainval, and the final trained model is submitted to online testing server for evaluation.

\renewcommand{\arraystretch}{1.5}
\begin{table*}[t]
	\centering
	\caption{Object detection results on PASCAL VOC 2012 \texttt{test} set. The performance is measured by mean of Average Precision (mAP, in \%).
			 }
	\label{table:VOC12}
	\resizebox{\textwidth}{!}{
    \setlength\tabcolsep{2.5pt}
	\begin{tabular}{c|c|c|cccccccccccccccccccc}
		\toprule 
		\textbf{Two-Stage} & Backbone
		& \textbf{mAP}
		& areo   & bike   & bird   & boat   & bottle & bus    & car    & cat    & chair  & cow
		& table  & dog    & horse  & mbk    & prsn   & plant  & sheep  & sofa   & train  & tv   \\
		\midrule 
		HyperNet~\cite{kong2016hypernet} & VGG16
		& 71.4
		& 84.2   & 78.5   & 73.6   & 55.6   & 53.7   & 78.7   & 79.8   & 87.7   & 49.6   & 74.9
		& 52.1   & 86.0   & 81.7   & 83.3   & 81.8   & 48.6   & 73.5   & 59.4   & 79.9   & 65.7 \\
		Faster~\cite{ren2015faster} & ResNet101
		& 73.8
		& 86.5   & 81.6   & 77.2   & 58.0   & 51.0   & 78.6   & 76.6   & 93.2   & 48.6   & 80.4
		& 59.0   & 92.1   & 85.3   & 84.8   & 80.7   & 48.1   & 77.3   & 66.5   & 84.7   & 65.6 \\
		ION~\cite{bell2016inside} & VGG15
		& 76.4
		& 87.5   & 84.7   & 76.8   & 63.8   & 58.3   & 82.6   & 79.0   & 90.9   & 57.8   & 82.0
		& 64.7   & 88.9   & 86.5   & 84.7   & 82.3   & 51.4   & 78.2   & 69.2   & 85.2   & 73.5 \\
		R-FCN~\cite{dai2016r}& ResNet101
		& 77.6
		& 86.9   & 83.4   & 81.5   & 63.8   & 62.4   & 81.6   & 81.1   & 93.1   & 58.0   & 83.8
		& 60.8   & 92.7   & 86.0   & 84.6   & 84.4   & 59.0   & 80.8   & 68.6   & 86.1   & 72.9 \\
		\bottomrule 
		\toprule 
		\textbf{One-Stage} & Backbone
		& \textbf{mAP}
		& areo   & bike   & bird   & boat   & bottle & bus    & car    & cat    & chair  & cow
		& table  & dog    & horse  & mbk    & prsn   & plant  & sheep  & sofa   & train  & tv   \\
		\midrule 
		SSD300*~\cite{liu2016ssd} & VGG16
		& 75.8
		& 88.1   & 82.9   & 74.4   & 61.9   & 47.6   & 82.7   & 78.8   & 91.5   & 58.1   & 80.0
		& 64.1   & 89.4   & 85.7   & 85.5   & 82.6   & 50.2   & 79.8   & 73.6   & 86.6   & 72.1 \\
		DSSD 321~\cite{fu2017dssd} & ResNet-101
		& 76.3
		& 87.3   & 83.3    & 75.4  &\bf{64.6}& 46.8 & 82.7    & 76.5  &\bf{92.9}& 59.5  & 78.3
		& 64.3   &\bf{91.5}& 86.6  & 86.6    & 82.1 &\bf{53.3}& 79.6  &\bf{75.7}& 85.2  &\bf{73.9}\\
		StairNet~\cite{woo2017stairnet} & VGG16
		& 76.4
		& 87.7    & 83.1   & 74.6   & 64.2   &\bf{51.3}& 83.6   & 78.0  & 92.0   & 58.9   &\bf{81.8}
		&\bf{66.2}& 89.6   & 86.0   & 84.9   & 82.6   & 50.9   & 80.5   & 71.8   & 86.2   & 73.5 \\
		\bf{WeaveNet} & VGG16
		&\bf{77.0}
		&\bf{88.5}&\bf{83.6}&\bf{76.6}& 63.0    & 51.2    &\bf{83.7}&\bf{79.6}& 92.7 &\bf{60.9}& 81.5
		& 65.2    & 90.4    &\bf{87.9}&\bf{87.0}&\bf{82.8}& 50.5    &\bf{80.8}& 73.0 &\bf{86.8}&\bf{73.9}\\
		\bottomrule 
	\end{tabular}
    }
\end{table*}
\renewcommand{\arraystretch}{1.2}


The evaluation results are summarized in Table~\ref{table:VOC12}. As can be seen from the results, our proposed method surpasses the competitors under the same backbone network with a large margin. In particular, a Reduced VGG16 base WeaveNet surpass the vanilla SSD by $1.2\%$ and improves the overall mAP by $0.6\%$ upon the one of the strongest multi-scale method --- StairNet~\cite{woo2017stairnet}.

We  also visualize  some detection results on the testing set  in Figure~\ref{fig:VOC12_viz}. Compared with the vanilla SSD, WeaveNet shows stronger ability for detecting both tiny and difficult objects. For the objects with medium sizes, WeaveNet   provides more accurate bounding boxes which fit the objects  tightly   thanks to its unique iterative information weaving.

\subsection{Conclusion}
In this work, we observe that both fine information from lower layers and coarse information from higher layers are crucial for building a highly efficient object detector. We propose a novel multi-scale fusion architecture, named WeaveNet.  WeaveNet iteratively \textit{weaves} information from adjacent scales, which not only gradually increases the detector receptive filed but also smoothly introduces more fine details from lower layers for robust and precise bounding box prediction. It can be easily trained  and deployed without  batch normalization and consumes very little additional computational cost, which make it  superior over existing multi-scale fusion methods. The experimental results well demonstrate the remarkable speed and accuracy advantages of the proposed WeaveNet on PASCAL VOC 2007 and PASCAL VOC 2012 dataset. In the near further, we would like to further evaluate the proposed WeaveNet on challenging MS COCO benchmark~\cite{lin2014microsoft}, and release all the trained models and source codes on GitHub.

{\small
\bibliographystyle{ieee}
\bibliography{egbib}
}

\end{document}